\documentclass[letterpaper, 10 pt, conference]{ieeeconf}

\IEEEoverridecommandlockouts

\newcommand{\AVAR}{\operatorname{AV@R}}  
\newcommand{\VAR}{\operatorname{V@R}}  

\usepackage{graphicx}
\usepackage{times}
\usepackage{graphicx}
\usepackage{amsmath}
\usepackage{amssymb}
\usepackage{cite}
\usepackage{hyperref}
\usepackage{algorithmic}

\usepackage{subcaption}
\usepackage{multirow}
\usepackage{xcolor}
\usepackage{booktabs}
\usepackage{makecell}
\usepackage{mathtools}
\usepackage{bm}
\usepackage{physics}
\usepackage{subcaption}
\usepackage{url}
\usepackage[linesnumbered,ruled,vlined]{algorithm2e}
\usepackage[utf8]{inputenc}
\usepackage[normalem]{ulem}
\usepackage{xcolor}
\usepackage{tikz}
\usetikzlibrary{arrows.meta,positioning,calc,fit,shapes.geometric,shapes.multipart,shapes.symbols}
\usetikzlibrary{backgrounds}

\usepackage{caption}
\newtheorem{theorem}{Theorem}

 \newcommand{\R}{{\mathbb{R}}}

\newcommand{\by}{\boldsymbol{y}}

\newcommand{\bw}{\boldsymbol{w}}

\def\by{{\bf \Phi}}

\def\bw{{\bf w}}

\def\bH{{\mathbf H}}

\usepackage{accents}
\newlength{\dhatheight}

\usepackage{pgf,tikz}

\usepackage[margin=1in]{geometry}

\begin{document}

\title{\LARGE \bf Active Next-Best-View Optimization for Risk-Averse Path Planning}



\author{
Amirhossein Mollaei Khass, Guangyi Liu, Vivek Pandey, Wen Jiang, Boshu Lei,  Kostas Daniilidis, \\ and Nader Motee%
\thanks{A.M. Khass, V. Pandey and N. Motee are with the Department of Mechanical Engineering and Mechanics, Lehigh University. {\tt\small \{ammb23, vkp219, motee\}@lehigh.edu}.}%
\thanks{W. Jiang, B. Lei, and K. Daniilidis are with the Department of Computer and Information Science, University of Pennsylvania, Philadelphia, PA 19104, USA. {\tt\small \{wenjiang, leiboshu\}@seas.upenn.edu}, {\tt\small kostas@cis.upenn.edu}.}%
}

\maketitle

\thispagestyle{empty}
\pagestyle{empty}

\begin{abstract}
Safe navigation in uncertain environments requires planning methods that integrate risk aversion with active perception. In this work, we present a unified framework that refines a coarse reference path by constructing tail-sensitive risk maps from Average Value-at-Risk statistics on an online-updated 3D Gaussian-splat Radiance Field. These maps enable the generation of locally safe and feasible trajectories. In parallel, we formulate Next-Best-View (NBV) selection as an optimization problem on the $\mathrm{SE}(3)$ pose manifold, where Riemannian gradient descent maximizes an expected information gain objective to reduce uncertainty most critical for imminent motion. Our approach advances the state-of-the-art by  coupling risk-averse path refinement with NBV planning, while introducing scalable gradient decompositions that support efficient online updates in complex environments. We demonstrate the effectiveness of the proposed framework through extensive computational studies.
\end{abstract}

\vspace{-0.0cm}



\section{Introduction}
Autonomous navigation in unknown or partially observable environments remains a fundamental challenge in robotics, particularly for applications requiring real-time mapping, localization, and risk-averse trajectory planning. In such settings, a robot must simultaneously build a model of its environment, localize itself within that model, and plan safe, goal-directed paths that account for uncertainty and potential hazards~\cite{Liu2024_RaEM, splatnav2025chen, chen2024safer-splat, Jiang2024_AGSLAM}. At the same time, it must actively acquire new observations to refine its internal map and improve decision-making.

While traditional SLAM techniques have made significant strides~\cite{chen2020active, carrillo2012comparison}, many systems—such as AG-SLAM~\cite{Jiang2024_AGSLAM}—treat {exploration}, {perception}, and {safe planning} as decoupled components. Active approaches typically focus on maximizing visibility or reducing global map ~\cite{rodriguez2018importance,matsuki2024gaussian}, without explicitly considering downstream task objectives like reaching a specific goal under safety constraints.

Recent advancements in {3D Gaussian Splatting (3DGS)}~\cite{Kerbl2023_3DGS} have introduced promising tools for efficient and high-fidelity scene reconstruction. These have been integrated into SLAM pipelines~\cite{Matsuki2024_GSSLAM} and extended to active perception frameworks like AG-SLAM~\cite{Jiang2024_AGSLAM}, which combines 3DGS with next-best-view strategies. However, such methods predominantly support {unconstrained exploration} and do not address the scenario where a robot must {navigate purposefully} toward a goal.

In contrast, safety-aware methods such as RaEM~\cite{Liu2024_RaEM} emphasize risk-aware viewpoint selection but often rely on predefined viewpoint sets and do not integrate real-time motion planning with onboard sensory feedback. Recent work such as Splat-Nav \cite{splatnav2025chen} leverages pretrained Gaussian Splatting scene representations to plan safe navigation corridors. 

Despite progress, a key gap remains: in realistic missions, robots are frequently required to follow a task-specific trajectory from a start to a goal position—where significant deviations from this path are undesirable or infeasible. Existing view planning and exploration algorithms are not well suited to this constrained setting, where safety must be ensured {along a predefined or planned route}, not in free space. This motivates our formulation, which tightly integrates information gathering with {trajectory-aware, risk-averse planning}.

In this work, we present a unified framework for risk-averse, perception-driven trajectory planning that tightly couples safe motion with informative sensing in unknown 3-D environments. A coarse reference path is refined by constructing conservative risk maps using Average Value-at-Risk ($\AVAR$) over an online-updated 3D Gaussian-splat Radiance Field, from which safe lattice points are extracted and searched with A*. Concurrently, next-best-views are optimized on the $\mathrm{SE}(3)$ pose manifold via Riemannian gradient descent on a Fisher-information objective to reduce uncertainty most critical for imminent motion. Our key contributions are: (i) a risk-averse replanning framework that integrates local A* with real-time $\AVAR$-based safety filtering using 3D Gaussian Splat Radiance Fields; (ii) a Riemannian optimization scheme on $\mathrm{SE}(3)$ for next-best-view computation, maximizing information gain under geometric and task-specific heuristics; and (iii) computational enhancements such as decomposable gradients that enable stochastic or mini-batch updates for scalability.

\begin{figure*}[t]
    \centering
    \includegraphics[width=\textwidth]{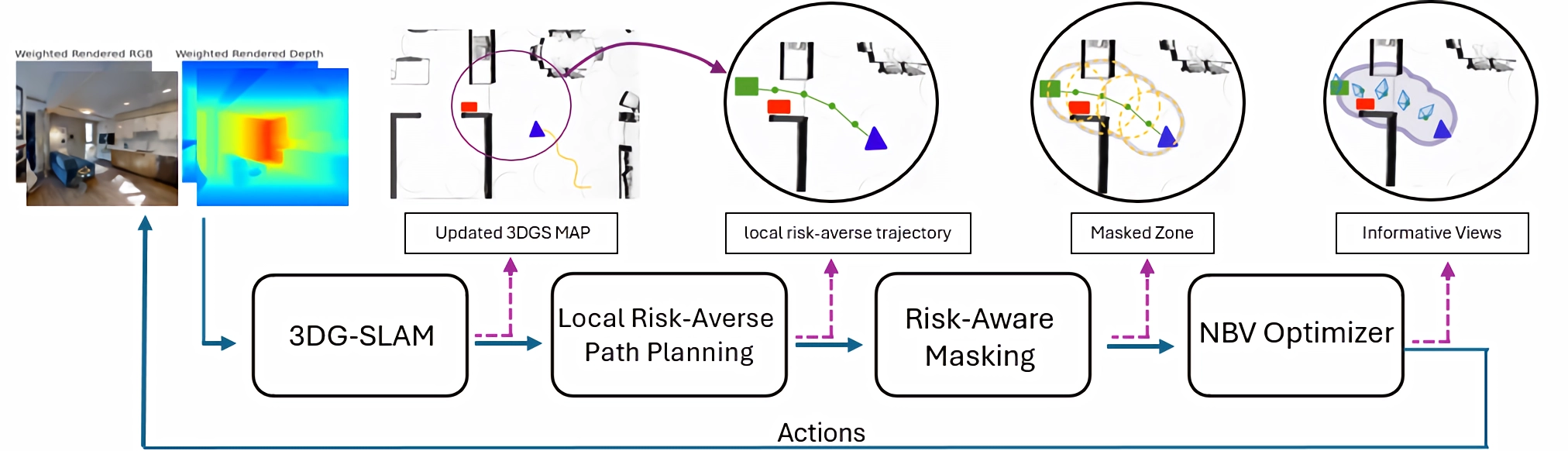}
\caption{{Block diagram of the proposed risk-averse trajectory planning as in Algorithm ~\ref{alg:trajectory-planning}. Each cycle the system updates the 3D Gaussian point cloud via SLAM, computes a risk field, synthesizes a safe local path, masks the environment around that path, selects high-risk splats, and evaluates candidate NBV poses over this focused subset. Executing the control and sensing actions steers the robot toward its goal while simultaneously gathering informative observations.
 } }
    \label{fig:pipeline}
    \vspace{-0.5cm}
\end{figure*}

\vspace{0.25cm}
\noindent
\section{Related Works}
Safe navigation in visually rich environments has gained renewed interest with the emergence of efficient scene representations like Gaussian Splatting (GSplat). These representations enable real-time rendering and dense mapping, making them well-suited for onboard planning and control. Splat-Nav~\cite{splatnav2025chen} introduces a two-part framework for safe navigation using pretrained GSplat scenes: Splat-Plan, which computes safe corridors via a fast ellipsoid-ellipsoid collision test, and Splat-Loc, which performs robust localization using RGB images through a PnP algorithm. Although their method could, in principle, be applied to dynamic scenes, the authors evaluate Splat-Nav only in preconstructed static environments, without real-time updates of the 3DGS, noting that existing GSplat SLAM algorithms do not yet run in real time. On the other hand, AG-SLAM \cite{Jiang2024_AGSLAM} extends GSplat to active SLAM, planning trajectories that balance information gain and localization uncertainty using Fisher Information. Unlike earlier exploration methods that assume perfect tracking, AG-SLAM accounts for pose uncertainty and improves robustness in challenging conditions. Still, it does not explicitly enforce safety during navigation. SAFER-Splat \cite{chen2024safer-splat} proposes a real-time safety filter based on a novel Control Barrier Function (CBF) that operates directly on GSplat primitives. SAFER-Splat assumes that the underlying GSplat map is already reasonably accurate and does not leverage active perception to update the environment. In contrast, our approach employs risk-aware safety measures and actively selects next-best-views to maximize expected information gain, enabling continuous map refinement in the vicinity of the robot’s trajectory during exploration. Due to space limitations, we restrict our discussion to recent works only.

\section{Problem Formulation}\label{sec:problem}

We consider an autonomous robot equipped with a forward-facing RGB-D camera and onboard computation that must traverse an unknown or partially known 3-D environment under sensing, safety, and resource constraints. The robot is provided a coarse reference trajectory and maintains an online 3D Gaussian-splat (3DGS) map updated from acquired views. The objectives are to (i) ensure collision-free operation, (ii) produce a near-optimal short trajectory, and (iii) selectively learn only the scene regions required for safe navigation to limit computation and communication. To this end, the planner alternates short-horizon, risk-aware path optimization with NBV computation: each waypoint is treated as a local subgoal, a forward-looking risk mask is formed from the current 3DGS risk field, and a proximity-weighted Expected-Information Gain (EIG) objective is optimized to produce NBV poses that maximally reduce uncertainty relevant to imminent motion. Executing the resulting control and sensing actions for the local path segment and assimilating the new images closes the loop, yielding a real-time, information-driven strategy for safe navigation in complex 3-D scenes.

\section{Technical Preliminaries} \label{sec:tech_prelims}
In this section, we establish the technical foundations of our approach.


\subsection{Learning 3D Scenes via Gaussian Splatting}
Volumetric rendering \cite{kajiya1984ray} forms the foundation of modern 3D scene reconstruction approaches such as NeRF \cite{Mildenhall2020_NeRF}, Plenoxels \cite{FridovichKeil2022_Plenoxels}, and 3D Gaussian Splatting \cite{Kerbl2023_3DGS}. The discrete volumetric rendering equation in 3D Gaussian Splatting (3DGS) \cite{turkulainen2024dnsplatter} produces the pixel color  as $    \hat{C} = \sum_{i=1}^{N_s} T_i \,\rho_i \, \mathbf{c}_i,$ and $ T_i = \prod_{j=1}^{i-1} (1 - \rho_j)$, where $T_i$ is the accumulated transmittance at pixel location \(x_i \in \R^2\), $\rho_i$ is the blending coefficient of the $i$-th Gaussian, and $\mathbf{c}_i$ is its color. The blending coefficient is defined as $    \rho_i = \epsilon_i \exp\big( -\frac{1}{2} (x_i - \mu_i)^\top \Sigma_i^{-1} (x_i - \mu_i) \big)$, where $\epsilon_i$ is the opacity, and $\mu_i$, $\Sigma_i$ define the 2D projected position and shape of the Gaussian in screen space. Depth rendering is computed analogously. The discrete depth rendering equation to produce $\hat{D}$ is given by 
$\hat{D} = \sum_{i=1}^{N_s} T_i \,\rho_i \, d_i,$ where $d_i$ is the $z$-depth of the $i$-th Gaussian in camera space. The model is optimized jointly using RGB and depth supervision:
\begin{equation}
    \mathcal{L} = \sum_{i = 1}^{N_s} \big( \mathcal{L}_1(C, \hat{C}) + \psi \, \mathcal{L}_1(D, \hat{D}) \big),
\end{equation}
where $C$ and $D$ are the ground-truth RGB and depth, $\hat{C}$ and $\hat{D}$ are the rendered predictions, $\mathcal{L}_1$ is the $L_1$ loss, and $\psi \in [0,1)$ balances depth relative to RGB.

\subsection{Quantifying Collision Risk in 3D Gaussian Splatting}

Since our ultimate objective is to plan a risk-averse path through an uncertain environment while simultaneously updating the scene representation, we begin by introducing a principled measure of safety. In this work, we employ the Average Value-at-Risk, formally defined as follows.

Let \( (\Omega, \mathcal{F}, \mathbb{P}) \) denote the underlying probability space. For a continuous random variable \( y : \Omega \to \mathbb{R} \), the Average Value-at-Risk ($\AVAR$) at level \( \varepsilon \in (0, 1) \) is defined as conditional expected value \cite{RockafellarUryasev2002_CVaR, Sarykalin2008_VaRvsCVaR}:
\begin{equation}
    \AVAR_{\varepsilon(y)} := \mathbb{E}\left[\, y \;\middle|\; y < \VAR_\varepsilon(y) \,\right],
\end{equation}
\noindent where the corresponding Value-at-Risk ($\VAR$) is given by $\VAR_\varepsilon(y) := \inf\left\{ z \in \mathbb{R} \;\middle|\; \mathbb{P}(y < z) > \varepsilon \right\}$.

We aim to construct a risk field over the environment, quantifying collision risk at every point in space. We begin by discretizing the workspace into a rectilinear grid with vertices $q = q_{ijk} = (\alpha_i, \beta_j, \gamma_k) \in \mathbb{R}^3,$ and denote the full set of vertices by $ \mathcal{Q} $. The 3D environment itself is modeled as a radiance field represented by Gaussian splats, where each splat is  $x_i \sim \mathcal{N}(\mu_i, \Sigma_i)$  with the full collection $ \mathcal{X}_t = \{x_i\} $.  To evaluate the local risk at a vertex $ q $, we first measure its relationship to each Gaussian $ x_i $. This is captured through the signed distance:  
\begin{equation}
    d(q, x_i) := \left\langle x_i - q,\, \frac{\mu_i - q}{\|\mu_i - q\|_2} \right\rangle,
    \label{eq:def_signed_distance}
\end{equation}
where $ \langle \cdot, \cdot \rangle $ is the Euclidean inner product \cite{Liu2024_RaEM}, and $ \|\cdot\|_2 $ denotes the $2$-norm. Assuming isotropic covariance $ \Sigma_i = \sigma_i^2 I_3$, this signed distance itself becomes a Gaussian random variable $d(q, x_i) \sim \mathcal{N} \big( \|\mu_i - q\|_2,\, \sigma_i^2 \big)$. We then quantify safety in terms of the Average Value-at-Risk (AV@R) of this distance distribution, which measures the expected value of the worst $ \varepsilon $-fraction of outcomes:  
\begin{equation}
    \operatorname{AV@R}_\varepsilon(d) = \|\mu_i - q\|_2 - 
    \frac{\sigma_i}{\sqrt{2\pi}} \cdot 
    \frac{1}{\varepsilon \exp(\iota^2)},
    \label{eq:avar_def}
\end{equation}
where $ \iota = \operatorname{erf}^{-1}(2\varepsilon - 1) $, and $ \operatorname{erf}^{-1}(\cdot) $ is the inverse error function. Finally, to obtain a conservative measure of risk at each vertex $ q $, we take the minimum AV@R value across all splats:  
\begin{equation}
    \alpha(q) := \min_{x_i \in \mathcal{X}_t} \operatorname{AV@R}_\varepsilon\big( d(q, x_i) \big).
    \label{eq:risk_min}
\end{equation}
Repeating this computation across all vertices $ q \in \mathcal{Q} $ yields risk a scalar field, which we interpret as the risk field of the environment. This risk field serves as the foundation for risk-averse planning and the NBV optimization. 


\subsection{Risk-Aware Environment Masking for NBV}\label{subsec: NBV}
Traditional NBV methods evaluate a finite pool of candidate camera poses and select the pose that maximizes information gain over the entire scene \cite{Jiang2024_FisherRF}. In our setting, the camera is rigidly mounted to the robot, so the robot pose and the camera orientation together determine the observed view. Rather than optimizing a global information objective, we weight information by its relevance to future risk along the planned trajectory by prioritizing observations that most effectively reduce collision uncertainty in regions that will be traversed by the robot. This risk-weighted NBV criterion concentrates sensing resources on the regions that directly affect the robot’s near-term motions, deliberately de-emphasizing distant parts of the scene that do not influence imminent decisions. By prioritizing measurements inside the forward-looking mask, the planner ignores far-range scenes unless and until it becomes relevant, which yields safer, lower-latency, and more computationally efficient exploration.

We construct a risk-driven masked environment and restrict the NBV search to this focused region. 
Given a coarse reference trajectory between two consecutive subgoals, represented by a sequence of 3D waypoints 
$\mathcal{Z}=\{z_1,z_2,\dots,z_N\}$ with $z_i\in\mathbb R^3$, each waypoint $z_k$ is assigned a masking radius  
\[
r_{\mathrm{mask}}(z_k)=\beta_1 e^{-\beta_2 \alpha(z_k)}, \quad \beta_1,\beta_2>0,
\]  
where $\alpha(z_k)$ denotes the risk field evaluated at $z_k$. The forward-looking masked region around $\mathcal{Z}$ 
is defined as the union of Euclidean balls  
\begin{equation}
\Pi_{\mathcal Z}=\bigcup_{z_k\in\mathcal Z}\mathcal B\bigl(z_k,r_{\mathrm{mask}}(z_k)\bigr).
\label{eq:mask_union}
\end{equation}

\begin{figure}[t]
    \centering
        \centering
        \includegraphics[width=0.50\linewidth]{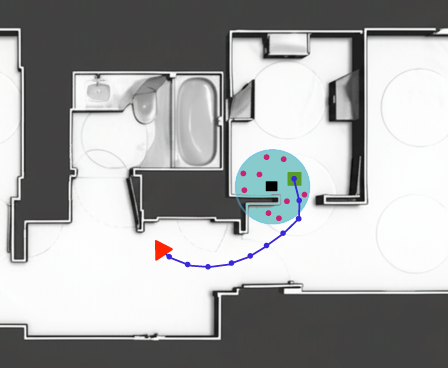}
        \caption{This shows a safe proxy subgoal selection as explained in Subsection \ref{subsec:V-A}. A safe subgoal \(\bar{z}_{j+1}\) (\textcolor{green}{\rule{2mm}{2mm}}) is chosen within a ball centered at the current unsafe subgoal \(z_{j+1}\) (\textcolor{black}{\rule{2mm}{2mm}}). Robot is depicted by \textcolor{red}{$\blacktriangle$} and the connected blue dots show the planned local trajectory \(\mathcal{Z}_j^s\). }
        \label{fig:proxy_subgoal}
    \label{fig:denmark_subgoal_navigavbtion}
    \vspace{-0.6cm}
\end{figure}

Thus, $\Pi_{\mathcal Z}$ collects the scene elements that most strongly influence the robot’s near-term collision risk; 
see Fig.~\ref{fig:denmark_subgoal_navigavbtion}.  

The next-best view is then selected by maximizing a proximity-weighted Fisher-information Expected Information Gain (EIG) 
over this masked region. Restricting the NBV search to $\Pi_{\mathcal Z}$ concentrates sensing and model updates on areas 
that directly affect imminent motion, thereby sparsifying acquisitions and reducing computation compared with global strategies.

\section{Optimizing NBV for Risk-Averse Planning} \label{sec:main_results}

Our proposed framework is depicted in Fig.~\ref{fig:pipeline} and summarized in Algorithm~\ref{alg:trajectory-planning}. Building upon the concepts introduced in Subsection~\ref{subsec: NBV}, we establish the foundations of our method and provide formal safety guarantees. Consider a robot deployed in an unknown and uncertain environment, tasked with following a coarse reference trajectory $\mathcal{Z} = \big\{z_1, z_2, \dots, z_N \big\}$. Since no safety assurances are inherent to this reference path, some waypoints may fall within unsafe regions. To address this challenge in real-time safety-critical navigation, we propose an {\it adaptive, localized, risk-averse} trajectory planning scheme. As illustrated in Fig.~\ref{fig:proxy_subgoal}, the scheme adjusts the next immediate waypoint (subgoal) to guarantee safety, while simultaneously identifying the most informative regions of the environment along the reference trajectory. These regions are used to update the 3DGS-based scene model, ensuring that the resulting trajectory remains provably safe.       


\subsection{Safety Verification for Local Trajectory Planning}\label{subsec:V-A}
In dynamically changing or partially known environments, the reference trajectory may not guarantee global safety. 
To ensure safe execution, we adopt a local replanning strategy guided by the localized risk profile. 
When the robot is located at waypoint $ z_j \in \mathcal{Z} $, the subsequent waypoint $ z_{j+1} $ is treated as a {\it subgoal}. 
The task is then to compute a risk-averse path from $ z_j $ to $ z_{j+1} $ using only local environmental information.  

To this end, we define a localized partition of the workspace, $ \mathcal{Q}_j \subseteq \mathcal{Q} $, which contains all grid points between $ z_j $ and $ z_{j+1} $. 
For each grid point $ q \in \mathcal{Q}_j $, the worst-case risk metric \eqref{eq:risk_min} is evaluated. 
The subset of safe candidate points is then obtained as  
\begin{equation}
    \mathcal{Q}^f_j = \left\{ q \in \mathcal{Q}_j \;\middle|\; \alpha_q \geq \gamma \right\},
    \label{eqn:filtered_Space}
\end{equation}
where $ \gamma \in \mathbb{R}_{+} $ denotes a user-specified risk tolerance. 
A collision-free, risk-averse path from $ z_j $ to $ z_{j+1} $ is computed over the filtered grid $ \mathcal{Q}^f_j $ using the standard $ A^\star $ algorithm.  

If the nominal waypoint $ z_{j+1} $ lies outside the safe set $ \mathcal{Q}^f_j $, it is refined by selecting the safest feasible alternative within a neighborhood of $ z_{j+1} $. 
The { proxy subgoal} is formally defined as  
\begin{equation} \label{eqn:proxy_goal}
    \bar{z}_{j+1} := \underset{z \in \mathcal{Q}^f_j \cap \mathcal{B}(z_{j+1}, \delta(z_{j+1}))}{\arg\mathrm{maximize}} \; \operatorname{AV@R}(z),
\end{equation}
where $ \mathcal{B}\bigl(z_{j+1}, \delta(z_{j+1}) \bigr) $ denotes the ball of radius $ \delta(z_{j+1}) $ centered at $ z_{j+1} $. 
In rare cases where no feasible { proxy subgoal} exists, the robot continues to explore within the safe region $ \mathcal{Q}^f_j $ until a suitable goal can be identified. 
An illustration of this procedure is provided in Fig.~\ref{fig:proxy_subgoal}.  

The locally replanned, risk-averse trajectory segment from $ z_j $ to either the refined proxy { subgoal} $ \bar{z}_{j+1} $ or the nominal waypoint $ z_{j+1} $ (when feasible) is denoted as  
\begin{equation}
    \mathcal{Z}_j^s = \{ r_j^1, r_j^2, \dots, r_j^{K_j} \},
\end{equation}
with $ r_j^1 = z_j $, $ r_j^{K_j} = z_{j+1}$ or $\bar{z}_{j+1}$, and $ K_j $ the number of points in the locally planned segment.

\begin{figure}[t]
    \centering
    \begin{subfigure}[b]{0.48\linewidth}
        \centering
        \includegraphics[width=\linewidth]{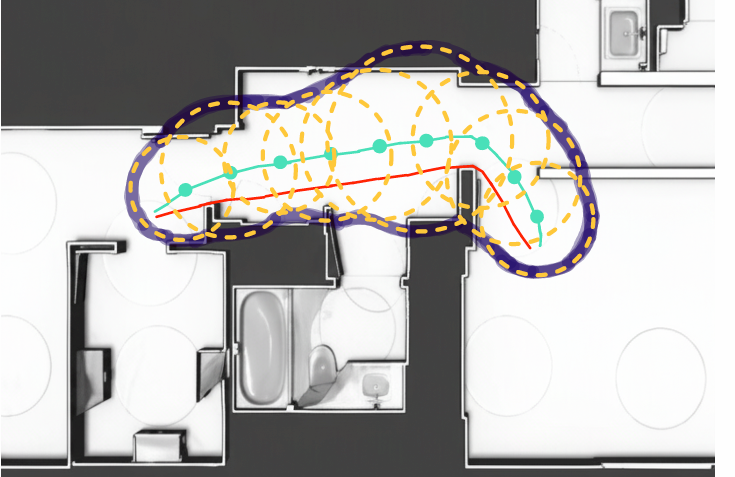}
        \caption{}
        \label{fig:safezone_mask}
    \end{subfigure}
    \hfill
    \begin{subfigure}[b]{0.48\linewidth}
        \centering
        \includegraphics[width=\linewidth]{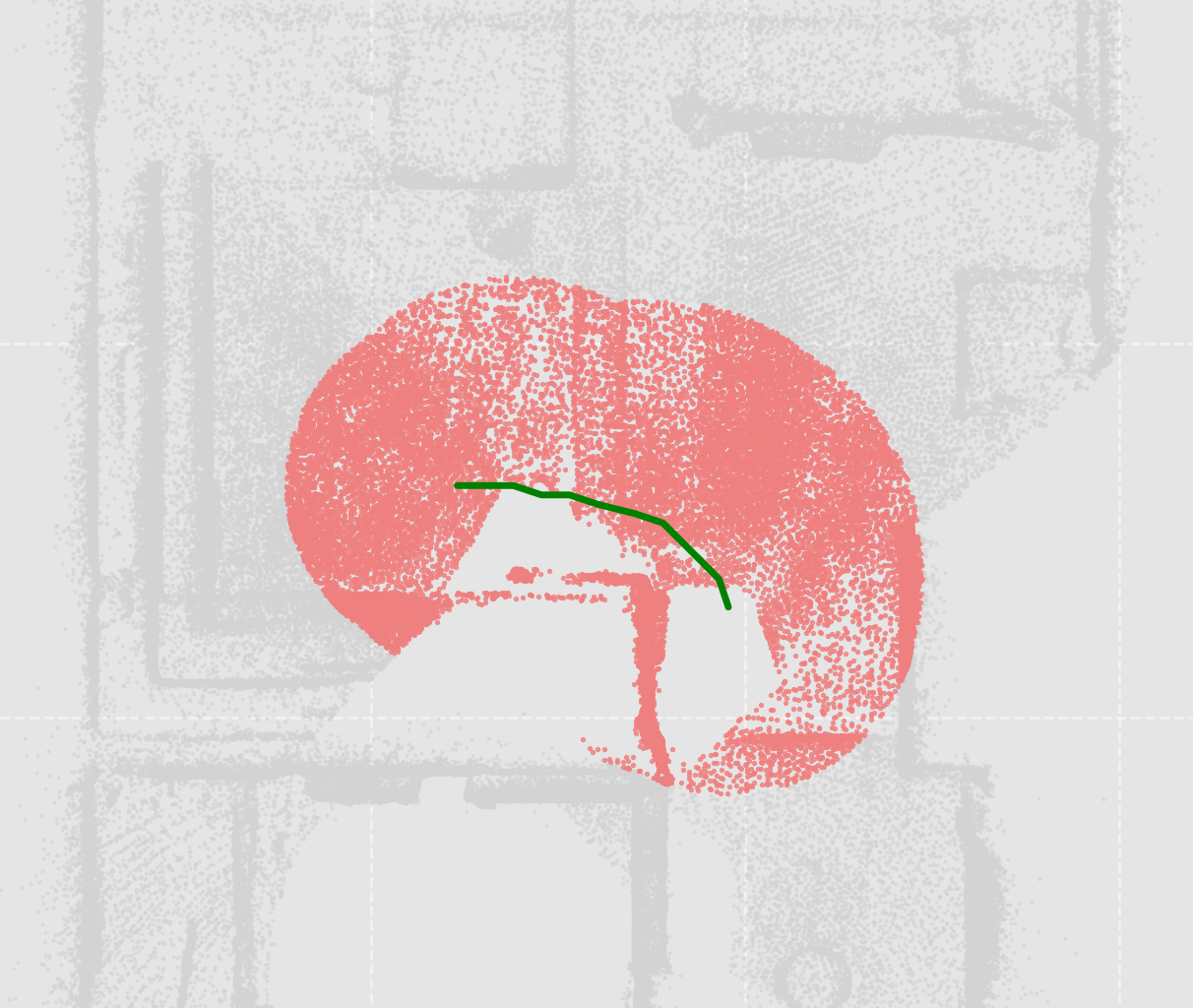}
        \caption{}
        \label{fig:gaussians_in_zone}
    \end{subfigure}
    \caption{
 Illustration of risk-aware environment masking to guide NBV selection, focusing on locally relevant Gaussians to maximize expected information gain and improve map accuracy. (a) Schematic of the masking process, showing Gaussian points near the trajectory in the masked region. (b) Subset of 3D Gaussian points selected within each safe zone to maximize information gain.
} 
    \label{fig:safezone_gaussian_selection}
    \vspace{-0.6cm}
\end{figure}

\subsection{NBV Optimization via Proximity-Weighted EIG}

Classical NBV planning strategies~\cite{Jiang2024_FisherRF,Wilson2025_POpGS} are effective for maximizing Expected Information Gain (EIG) in unconstrained exploration. 
However, they are less suitable when the robot must follow a prescribed trajectory from start to goal while adhering to safety constraints. 
In such task-driven scenarios, the robot cannot deviate significantly from its intended path.  

To address this limitation, we propose a trajectory-aware NBV strategy that builds on the principles in Subsection~\ref{subsec: NBV} and operates along the locally safe trajectory segment $ \mathcal{Z}_j^s $. 
At each waypoint $r_j^k$, the robot selects the most informative view, thereby balancing information acquisition with path feasibility. Formally, the NBV problem at each \(r_j^k \in \mathcal{Z}_j^s\) can be written as 
\begin{equation}\label{eq:opt_Pi_t}
T_{k,j}^\ast =\arg  \underset{T_{k,j}\in\mathrm{SE}(3)}{\mathrm{maximize}} ~ \mathcal I\bigl(T_{k,j};\Pi_{{\mathcal{Z}}_j^s}\bigr),
\end{equation}
where $\mathcal{I}\bigl(T_{k,j};\Pi_{{\mathcal{Z}}_j^s}\bigr)$ 
is the EIG as a function of the robot pose $T_{k,j}$ at waypoint \(r_j^k\), parameterized by the 3D Gaussian points in $\Pi_{{\mathcal{Z}}_j^s}$. Following~\cite{Jiang2024_FisherRF}, the information gain for the masked region $\Pi_{\mathbf{\mathcal{Z}_j^s}} $ is expressed as  
\begin{equation} \label{eqn:information_gain}
    \mathcal{I}\!\bigl(T; \Pi_{\mathbf{\mathcal{Z}_j^s}}\bigr) 
    = \operatorname{tr} \bigl( \bH''[\by \mid T, \bw^*] \; 
    \bH''[\bw^*]_{\text{prior}}^{-1} \bigr),
\end{equation}
where $\operatorname{tr}(Y)$ denotes the trace of matrix $Y$, and $\bH''[\bw^*]_{\text{prior}}^{-1}$ is obtained by aggregating Hessians of the 3DGS model parameters across all training views. 
The Hessian contribution of a candidate view is approximated as  
\begin{equation}    \label{eq:hessian_new_view}
        \bH''[\by \mid T, \bw^*] \;\approx\; 
        \operatorname{diag}\!\big(J^\top J\big) + \lambda I,
\end{equation}
where $J = \!\nabla_\bw f(T, \bw^*)$ is the Jacobian of the rendering model $f(T,\bw)$ with respect to Gaussian splat parameters $\bw$, $\operatorname{diag}\big(J^\top J\big)$ extracts the diagonal of $J^\top J$, and $\lambda$ is a regularization constant.  

To capture spatial relevance, we introduce a {\it proximity weighting function} that emphasizes nearby Gaussian splats relative to distant ones in $\Pi_{\mathbf{\mathcal{Z}_j^s}}$. 
This reflects the intuition that nearby Gaussian splats carry greater importance for safe navigation, since the robot encounters them before more distant structures. 
We define an exponential decay weight function with respect to the graph distance between the robot and splat center:
\begin{equation}\label{eqn:V_function}
    v_T(\mu_i) = \alpha \hspace{0.05cm} e^{-\beta \left\|p_{T} - \mu_i\right\|_g},
\end{equation}
where $\alpha,\beta > 0$ are design parameters, $ \left\|\cdot\right\|_g $ denotes graph (grid-based) distance, $p_T$ is the camera position from pose $T$, and $ \mu_i $ is the mean of the $i$th Gaussian splat.

\begin{algorithm}[t]
\caption{Risk-Averse Trajectory Planning}
\label{alg:trajectory-planning}
\begin{algorithmic}
    \STATE \textbf{Initialize:} Coarse reference trajectory from start to goal \(\mathcal{Z} =\{z_1, \dots, z_N\}\)
    \FOR{\(j = 1:N\)}
        \STATE Find safe filtered grid \(\mathcal{Q}_j^f = \left\{ q \in \mathcal{Q}_j \;\middle|\; \alpha(q) \geq \gamma \right\}\)
        \IF{\( z_{j+1} \notin \mathcal{Q}^f_j \)}
            \STATE Find safe proxy subgoal \(\bar{z}_{j+1}\) of \({z}_{j+1}\) 
\begin{equation*}
    \bar{z}_{j+1} :=  \underset{z \in \mathcal{Q}^f_j \cap \mathcal{B}(z_{j+1}, \delta(z_{j+1}))}{\arg\mathrm{maximize}} \; \AVAR(z),
\end{equation*}
        \ENDIF
        \STATE Find shortest path \( \mathcal{Z}_j^s \) from \(z_j\) to \(\bar{z}_{j+1}\) using \(A^{\star}\)
        Find risk-aware masked environment \(\Pi_{\mathbf{\mathcal{Z}_j^s}} \) 
        \FOR{\(k = 1:K_j\)}
        \STATE Find NBV for \(r_j^k \in \mathcal{Z}_j^s\)
        \begin{equation*}
    \psi^* =  \arg \underset{\psi \in \mathbb{S}^1}{\mathrm{maximize}} ~  \bar{\mathcal{I}}\big(T; \Pi_{\mathbf{\mathcal{Z}_j^s}}\big),
\end{equation*}
\vspace{-0.3cm}
         \STATE Update 3D Gaussian Splatting
        \ENDFOR
    \ENDFOR
    \STATE \textbf{Return:} Safe Path and Updated 3D Gaussian Map
\end{algorithmic}
\end{algorithm}

The weighting function is incorporated into the Fisher Information Matrix by modifying the Hessian approximation in \eqref{eq:hessian_new_view}:  
\begin{equation}\label{eqn:modified_hessian_next_view}
    \bar{\bH}''[\by \mid T, \bw^*] \approx 
   \left[ \bar{V} \operatorname{diag}\left(J^\top J\right) + \lambda I\right],
\end{equation}
where $\bar{V} = \operatorname{diag}(V_1,\dots,V_{|\mathcal{X}_T|})$, with $V_i = v_T(\mu_i)\cdot I_{|w_i|}$, scales the contribution of each Gaussian according to its proximity to the viewpoint, and $w_i$ is the parameter vector for splat $x_i \in \mathcal{X}$. The resulting proximity-weighted information gain is defined as  
\begin{equation} \label{eqn:information_gain}
    \bar{\mathcal{I}}\bigl(T; \Pi_{\mathbf{\mathcal{Z}_j^s}}  \bigr) 
    = \operatorname{tr} \bigl( \bar{\bH}''[\by \mid T, \bw^*] \; 
    \bH''[\bw^*]_{\text{prior}}^{-1} \bigr),
\end{equation}
which can be expressed equivalently as  
\begin{equation}\label{eqn:information_gain_summation_Form}
    \bar{\mathcal{I}}(T; \Pi_{\mathbf{\mathcal{Z}_j^s}} ) 
    = \sum_{i=1}^{|\mathcal{X}|} v_T(\mu_i) \, \mathcal{I}_i\!\bigl(T; \Pi_{\mathbf{\mathcal{Z}_j^s}}\bigr),
\end{equation}
where $ \mathcal{I}_i\!\bigl(T; \Pi_{\mathcal{Z}_j^s} \bigr) = \operatorname{tr}\!\big(\bar{\bH}''[\by \mid T, w_i^*] \, \bH''[w_i^*]_{\text{prior}}^{-1}\big) $ denotes the information gain associated with parameter $w_i$ and $|\mathcal{X}|$ is the cardinality of set $\mathcal{X}$. The full pose optimization in \eqref{eq:opt_Pi_t} considers all six degrees of freedom in $\mathrm{SE}(3)$. 
In the trajectory-aware NBV formulation, however, the robot is constrained to a fixed sequence of positions $\mathcal{Z}_j^s$. 
The translational component of $T$ is thus fixed, and the orientation is restricted to yaw rotations due to platform or sensor field-of-view constraints. 
This reduces the optimization to a single parameter, the yaw angle $\psi \in \mathbb{S}^1$:  
\begin{equation} \label{eq:opt_Pi_t_modified} 
    \arg \underset{\psi \in \mathbb{S}^1}{\mathrm{maximize}} \;\;  
    \bar{\mathcal{I}}\bigl(T; \Pi_{\mathbf{\mathcal{Z}_j^s}}  \bigr),
\end{equation}
where $\psi$ represents the viewing angle at each waypoint.  

We solve \eqref{eq:opt_Pi_t_modified} using gradient ascent on the manifold $\mathbb{S}^1$. 
The objective $\bar{\mathcal{I}}(\cdot)$ is differentiable with respect to $\psi$ under mild assumptions, enabling first-order methods. 

The process is repeated until convergence or until a maximum number of iterations is reached. 
In practice, only a few gradient steps are sufficient to yield an informative viewpoint policy. 

\begin{figure}[t]
    \centering
    \begin{subfigure}{0.21\textwidth}
        \includegraphics[width=\linewidth]{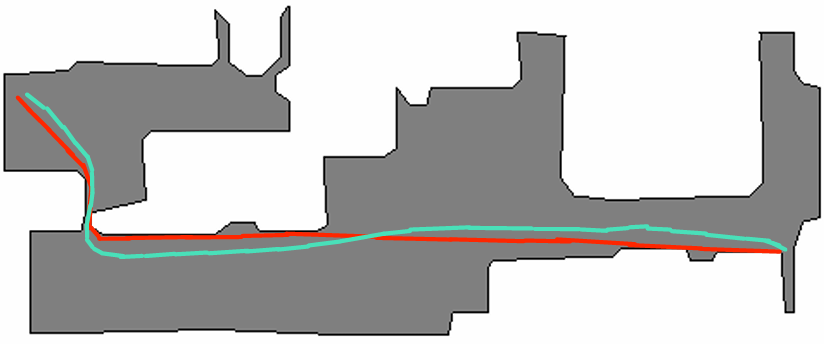}
        \caption{Pablo}
        \label{fig:subfig2}
    \end{subfigure}\hfill
    \begin{subfigure}{0.21\textwidth}
        \includegraphics[width=\linewidth]{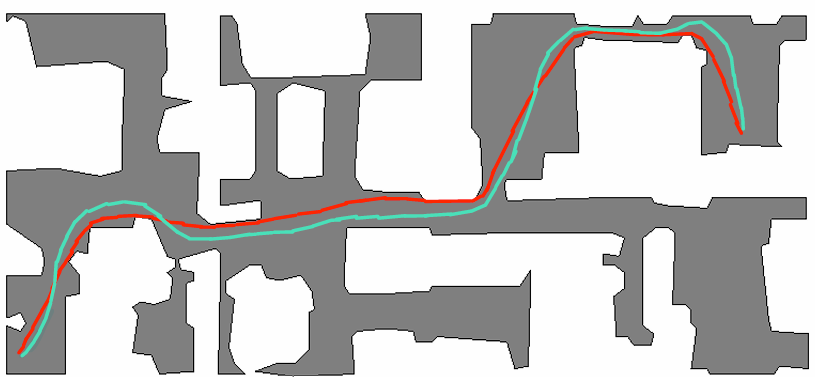}
        \caption{Cantwell}
        \label{fig:subfig3}
    \end{subfigure}\hfill

    \caption{Shortest path ($\textcolor{red}{\rule{4mm}{0.3mm}}$) and risk-averse path ($\textcolor{green}{\rule{4mm}{0.3mm}}$) shown in two representative environments; the risk-averse trajectory avoids high-risk regions at the expense of additional length.
}
    \label{fig:safe_trajectory_env}
    \vspace{-0.5cm}
\end{figure}

\subsection{Efficient Implementation of NBV Optimization}


While the information-theoretic optimization is effective, computing the full gradient at each iteration can be costly in large-scale environments. 
To improve runtime, we exploit the additive structure of the information gain objective in \eqref{eqn:information_gain_summation_Form}, reformulating \eqref{eq:opt_Pi_t_modified} as  
\begin{equation} \label{eq:opt_Pi_t_modified_summation} 
    \arg \underset{\psi \in \mathbb{S}^1}{\mathrm{maximize}} \;\;  \sum_{i =1}^{|\mathcal{X}|} v_{T}(\mu_i) \, \mathcal{I}_i\!\bigl(T; \Pi_{\mathbf{\mathcal{Z}_j^s}} \bigr).
\end{equation}

By linearity of the gradient operator, the update becomes a weighted sum of per-splat gradients:  
\begin{equation}\label{eqn:NBV_gradient_descent_summation}
    \psi^{(l+1)} \leftarrow \psi^{(l)} + \eta \sum_{i =1}^{|\mathcal{X}|} v_{T}(\mu_i) \nabla_\psi \mathcal{I}_i\!\bigl(T; \Pi_{\mathbf{\mathcal{Z}_j^s}}  \bigr).
\end{equation}
where $\eta > 0$ is the step size, and the gradient $\nabla_\psi \bar{\mathcal{I}}(\cdot)$ is obtained via automatic differentiation through the trajectory parameterization. This decomposition enables parallelization and supports efficient variants such as stochastic and mini-batch gradient ascent, which trade exactness for scalability and faster convergence, which is important for real-time and resource-constrained implementation.   

\begin{theorem}\label{thm:info_gain_ordering}
Consider two prior information matrices 
\(\bH''[\bw^*]_{\text{prior}}\) and \(\bH''[\mathbf{v}^*]_{\text{prior}},\) corresponding to parameters \(\bw^*\) and \(\mathbf{v}^*\) of the rendering model.  
If the prior satisfies the Loewner ordering on the cone of positive semi-definite matrices:
\[
\bH''[\bw^*]_{\text{prior}} \succeq \bH''[\mathbf{v}^*]_{\text{prior}},
\]
then the proximity-weighted EIG satisfies
\[
\bar{\mathcal{I}}_{\bw^*}\!\bigl(T; \Pi_{\mathcal{Z}_j^s}\bigr) \leq
\bar{\mathcal{I}}_{\mathbf{v}^*}\!\bigl(T; \Pi_{\mathcal{Z}_j^s}\bigr), 
\]
where 
\[
\bar{\mathcal{I}}_{\square}\!\bigl(T; \Pi_{\mathcal{Z}_j^s}\bigr) = 
\operatorname{tr} \bigl( \bar{\bH}''[\by \mid T, \square^*] \; 
    \bH''[\square^*]_{\text{prior}}^{-1} \bigr),
\] 
for \(\square \in \{\bw,\mathbf{v}\}.\) 
\end{theorem}


Theorem~\ref{thm:info_gain_ordering} (Proof in Appendix) shows that EIG exhibits diminishing returns: views added to a well-trained model yield less information than those added earlier. 
We exploit this as a principled early-stopping rule for \eqref{eqn:NBV_gradient_descent_summation}. 
Once the expected information gain falls below a threshold, further updates contribute negligibly to uncertainty reduction. 
Unlike fixed iteration limits or gradient-norm criteria, this stopping rule adapts to the informativeness of new views, ensuring efficient use of computation without sacrificing performance.

\section{EXPERIMENTS} \label{sec:sim_results}
In this section, we provide a brief overview of our experimental setup and discuss the results obtained from our algorithm \ref{alg:trajectory-planning}.


\begin{table*}[t]
\centering
\renewcommand{\arraystretch}{1.2}
\setlength{\tabcolsep}{6pt}
\begin{tabular}{l|ccc|ccc|ccc|ccc}
\toprule
\textbf{Environment} 
& \multicolumn{3}{c|}{\textbf{PSNR ↑}} 
& \multicolumn{3}{c|}{\textbf{SSIM ↑}} 
& \multicolumn{3}{c|}{\textbf{LPIPS ↓}} 
& \multicolumn{3}{c}{\textbf{ Depth MAE ↓}} \\
\cmidrule(lr){2-4} \cmidrule(lr){5-7} \cmidrule(lr){8-10} \cmidrule(lr){11-13}
& R=1 & R=2 & R=3 
& R=1 & R=2 & R=3 
& R=1 & R=2 & R=3  
& R=1 & R=2 & R=3  \\
\midrule
Denmark     &16.71 & 16.53 & 16.23 & 0.65  & 0.64 & 0.63 & 0.47 & 0.49  & 0.50 &  0.31& 0.32 & 0.34   \\
Pablo       &  15.79   &  15.40   &  14.20   &   0.64  &  0.62    &   0.60  &  0.47   &   0.48  & 0.51 & 0.43 & 0.48 & 0.53    \\
Cantwell    & 15.41 & 14.87 & 14.04 & 0.57  & 0.54 & 0.52 & 0.55 & 0.56  & 0.58 & 0.22 & 0.25 & 0.30   \\
Eudora      & 15.77 & 14.84 &  13.66  & 0.71 & 0.66 & 0.60 &  0.46 & 0.50 & 0.52 & 0.15 & 0.32 & 0.40   \\
Swormnville & 15.46 & 15.04 & 14.40 & 0.66  & 0.63 & 0.62 & 0.49 & 0.51  & 0.53 & 0.50 & 0.49 & 0.51   \\
Elmira      & 14.51 & 13.32 & 12.97 & 0.61  & 0.59 & 0.57 & 0.50 & 0.51  & 0.58 & 0.35 & 0.38 & 0.42   \\
\bottomrule
\end{tabular}
\caption{
Comparison of environment reconstruction metrics across 3 corridor radii (\(R = 1, 2, 3\),) around robot's trajectory. Higher PSNR and SSIM and lower LPIPS and MAE indicate better reconstruction quality. As expected, quality degrades with increasing distance from the robot's trajectory. 
}
\label{tab:metrics_corridor}
    \vspace{-0.5cm}
\end{table*}

\subsection{Experimental Setup}
The environment is simulated using the {Habitat} simulator~\cite{habitat19iccv} with scenes from the {Gibson} dataset~\cite{gibson}. All experiments are conducted on a machine with an {Intel Core i9-13900K CPU} and an {NVIDIA RTX A2000 GPU}, implemented in {PyTorch 2.2} with {CUDA 12}. Mask parameters are fixed to \((\beta_1, \beta_2) = (0.2, 1.1)\), with a tolerance of \(\gamma = 0.10\).

\begin{figure}[t]
    \centering
    \begin{subfigure}[b]{0.48\linewidth}
        \centering
        \includegraphics[width=\linewidth]{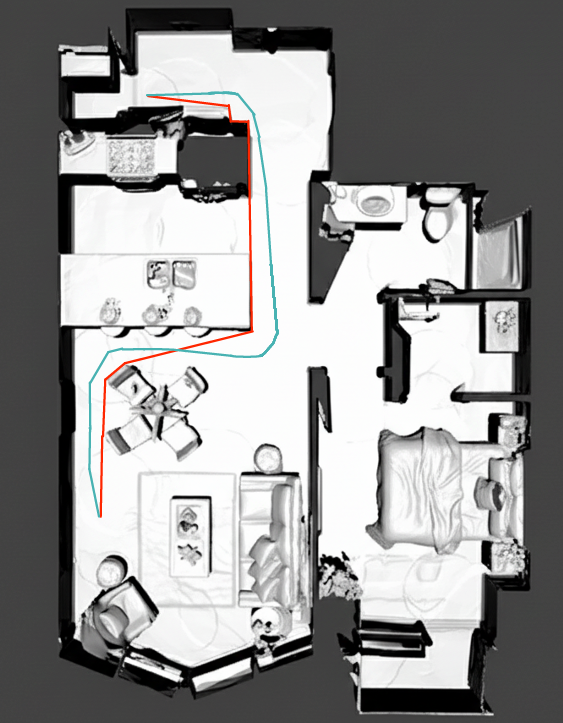}
        \caption{}
        \label{fig:swormnille_recon}
    \end{subfigure}
    \hfill
    \begin{subfigure}[b]{0.48\linewidth}
        \centering
        \includegraphics[width=\linewidth]{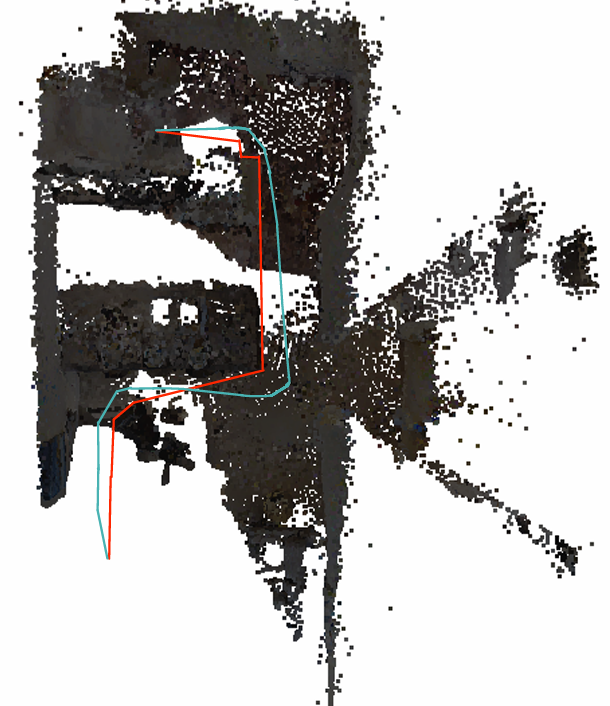}
        \caption{}
        \label{fig:swormnille_path}
    \end{subfigure}
    \caption{%
Results from the Swormville scene (Gibson dataset). (a) Top-down map showing the ground-truth shortest path ($\textcolor{red}{\rule{4mm}{0.3mm}}$) and the executed risk-averse trajectory ($\textcolor{green}{\rule{4mm}{0.3mm}}$). (b) 3-D Gaussian-splat reconstruction of the scene; only the forward-looking masked region relevant to the task was used for learning, illustrating the sparse, task-focused sensing.
}
    \label{fig:swormnille_results}
        \vspace{-0.6cm}
\end{figure}

\subsection{Evaluation Metrics} 




We evaluate our method across two key dimensions: trajectory safety and environment reconstruction quality.

To quantify {safety}, we use the {Average Value at Risk \(\operatorname{AV@R}\)} metric, which quantifies safety along a trajectory. A lower \(\AVAR\) indicates a higher concentration of risk, whereas a higher \(\AVAR\) implies that the robot is consistently avoiding high-risk regions. Thus, {higher \(\AVAR\) values reflect safer trajectories} in our context.

For assessing {environment reconstruction}, we employ standard image-based metrics: {PSNR (Peak Signal-to-Noise Ratio)} \cite{hore2010psnr}, {SSIM (Structural Similarity Index)} \cite{wang2004ssim}, {LPIPS (Learned Perceptual Image Patch Similarity)} \cite{Zhang2018_LPIPS}, and {MAE (Mean Absolute Error)}. These metrics evaluate how closely the rendered outputs from our Gaussian Splatting scene representation match ground-truth views, measuring both pixel-wise accuracy and perceptual quality.

While we also report {path length} for completeness, we note that safety-oriented planning naturally leads to longer trajectories compared to shortest-path baselines (e.g., \(A^\star\)). Therefore, path length is not used as a primary metric, but rather as a reference for analyzing the trade-off between safety and efficiency.

\subsection{Effectiveness of our Algorithm}


\subsubsection{Path Planning}

We evaluate our risk-aware planning algorithm across diverse environments, as illustrated in Fig.~\ref{fig:safe_trajectory_env}. The green trajectories represent paths generated by our method, while the red trajectories correspond to the shortest paths computed using \(A^{\star}\)\cite{hart1968a}.

As visually evident from the figures, our planner consistently steers the robot away from high-risk areas, maintaining greater clearance from obstacles compared to the shortest-path baseline. This qualitative observation highlights the effectiveness of our approach in producing safer, more robust trajectories.

Quantitatively, Table~\ref{tab:safe_trajectory} reports the path lengths and corresponding \(\operatorname{\AVAR}\) values. While our method results in longer trajectories, as expected in risk-averse navigation, it achieves significantly higher \(\AVAR\) scores, affirming its ability to minimize exposure to high-risk regions and prioritize safety over raw efficiency.





\subsubsection{Environment Reconstruction Quality}

We evaluate the fidelity of environment reconstruction using standard image quality metrics—PSNR, SSIM, LPIPS, and depth MAE—reported in Table~\ref{tab:metrics_corridor}. Our method employs a risk-aware masking strategy that emphasizes regions near the robot's trajectory when quantifying information gain. This prioritization ensures that areas critical for safe navigation are reconstructed with higher fidelity, aligning with our objective of localized, risk-sensitive mapping.

To further assess reconstruction performance, we evaluate the metrics within varying radii around the robot’s path, effectively forming a spherical corridor centered on the trajectory. As the radius increases, PSNR and SSIM values consistently decrease, while LPIPS and depth MAE values increase indicating that perceptual and geometric accuracy diminish in regions farther from the robot. This trend is not only expected, but also desirable, as it confirms that our algorithm focuses reconstruction quality where it matters most: in the immediate vicinity of the robot's trajectory. The consistent behavior across all metrics supports the effectiveness of our risk-aware strategy, and all observations are quantitatively supported by the results presented in Table~\ref{tab:metrics_corridor}.

We evaluate our NBV optimization (Eq.~\eqref{eqn:NBV_gradient_descent_summation}) by measuring the percentage increase in information gain achieved by the optimized view compared to a nominal one. As shown in Table~\ref{tab:metrics_reshaped}, the optimized NBV consistently yields higher information gain. While both the safety measure (Table~\ref{tab:safe_trajectory}) and expected information gain (EIG) (Table~\ref{tab:metrics_reshaped}) show an increasing trend, they are not strongly correlated, as EIG is influenced by the inherent complexity of each environment. 

\begin{table*}[t]
\centering
\renewcommand{\arraystretch}{1.2}
\setlength{\tabcolsep}{8pt}
\begin{tabular}{l|cccccc}
\toprule
\textbf{Environment} & \textbf{Denmark} & \textbf{Pablo} & \textbf{Cantwell} & \textbf{Eudora} & \textbf{Swormville} & \textbf{Elmira} \\
\midrule
\(\%\) EIG  \(\uparrow\)     & 41.6 & 28.0 & 35.2 & 20.21 & 33.60 & 20.83 \\

\bottomrule
\end{tabular}
\caption{
\(\%\) increase in Expected Information Gain (EIG) from optimized NBV compared to a nominal view. 
}
\label{tab:metrics_reshaped}
    \vspace{-0.5cm}
\end{table*}

\begin{table}[t]
\centering
\renewcommand{\arraystretch}{1}
\setlength{\tabcolsep}{3pt} 
\begin{tabular}{l|ccc|ccc}
\toprule
\textbf{Environment} 
& \multicolumn{3}{c|}{\textbf{Path Length} (m)} 
& \multicolumn{3}{c}{\textbf{Safety Measure} (m)} \(\uparrow\) \\
\cmidrule(lr){2-4} \cmidrule(lr){5-7}
& \textbf{Shortest} & \textbf{Ours} & \% loss
& \textbf{Shortest} & \textbf{Ours} & \% gain \\
\midrule
Denmark     & 5.72 & 7.79 & 36.1 & 0.27 & \textbf{0.37} & 37.0 \\
Pablo       & 5.81 & 6.68 & 15.0 & 0.82 & \textbf{0.96} & 17.1 \\
Cantwell    & 8.60 & 11.71 & 36.2 & 0.34 & \textbf{0.38} & 11.7 \\
Eudora      & 6.48 & 7.69 & 18.7 & 0.18 & \textbf{0.23} & 28.7 \\
Swormnville & 9.27 & 12.14 & 30.9 & 0.31 & \textbf{0.42} & 35.5 \\
Elmira      & 5.51 & 7.15 & 29.7 & 0.28 & \textbf{0.46} & 64.2 \\
\bottomrule
\end{tabular}
\caption{
Comparison of path lengths—Shortest (\(\textcolor{red}{\rule{4mm}{0.3mm}}
\)) as in Fig.~\ref{fig:swormnille_path}) and Ours (\(\textcolor{green}{\rule{4mm}{0.3mm}}
\)) as in Fig.~\ref{fig:swormnille_path}—and corresponding \(\AVAR\) values across environments.}
\label{tab:safe_trajectory}
    \vspace{-0.5cm}
\end{table}

\section{Conclusion}\label{sec:conclusion}
In this work, we tackled the problem of risk-averse navigation in unknown 3D environments. Our framework refines a coarse global path into a safe, perception-driven trajectory by maintaining an online 3D Gaussian Splatting (3DGS) map and actively reducing uncertainty through a NBV strategy optimized on the $\mathrm{SE}(3)$ manifold. By quantifying risk directly from the 3DGS representation, the robot can avoid unsafe regions while continuously replanning around local subgoals. This tight integration of mapping, risk assessment, $\mathrm{SE}(3)$-based viewpoint optimization, and planning enables navigation that is both safe and efficient, while preserving reconstruction quality where it matters most. Our experiments demonstrate that the framework delivers reliable, high-fidelity navigation in complex environments, highlighting its potential for real-world deployment.



\bibliographystyle{IEEEtran}
\bibliography{references}

\appendix


\noindent\textit{Proof of Theorem 1:} 
The result follows from the Loewner ordering of positive semidefinite matrices. For simplicity, let us define the Hessian corresponding to the new view as
\[\scalebox{0.95}{$
H_n = \bar{\bH}''[\by \mid T, \bw^*]^{1/2} = \bar{\bH}''[\by \mid T, \mathbf{v}^*]^{1/2}.
$}\]
Using the Loewner order, we have
\[\scalebox{0.95}{$
\bH''[\bw^*]_{\text{prior}} \succeq \bH''[\mathbf{v}^*]_{\text{prior}} 
 \implies 
\bH''[\bw^*]_{\text{prior}}^{-1} \preceq \bH''[\mathbf{v}^*]_{\text{prior}}^{-1}.
$}\]
Pre- and post-multiplying by $H_n^{1/2}$ preserves the order:
\[\scalebox{0.95}{$
H_n^{1/2}\, \bH''[\bw^*]_{\text{prior}}^{-1}\, H_n^{1/2} 
\preceq 
H_n^{1/2}\, \bH''[\mathbf{v}^*]_{\text{prior}}^{-1}\, H_n^{1/2}.
$}\]

Taking the trace on both sides yields
\begin{align*}
\operatorname{tr}\bigl(H_n^{1/2}\, \bH''[\bw^*]_{\text{prior}}^{-1}\, H_n^{1/2}\bigr) 
&\leq 
\operatorname{tr}\bigl(H_n^{1/2}\, \bH''[\mathbf{v}^*]_{\text{prior}}^{-1}\, H_n^{1/2}\bigr),\\[2mm]
\operatorname{tr}\bigl(H_n\, \bH''[\bw^*]_{\text{prior}}^{-1}\bigr) 
&\leq 
\operatorname{tr}\bigl(H_n\, \bH''[\mathbf{v}^*]_{\text{prior}}^{-1}\bigr),
\end{align*}
which establishes the claim. \hfill \(\blacksquare\)

\end{document}